\documentclass[conference]{IEEEtran}
\IEEEoverridecommandlockouts
\usepackage{cite}
\usepackage{amsmath,amssymb,amsfonts}
\usepackage{algorithmic}
\usepackage{graphicx}
\usepackage{textcomp}
\usepackage{xcolor}
\usepackage{dblfloatfix}
\usepackage{booktabs}
\usepackage{multirow}

\newcommand{\spotwo}{SpO$_2$\;}
\newcommand{\ie}{i.e.,\;}
\newcommand{\eg}{e.g.,\;}

\newcommand\SP[1]{}

\newcommand\revision[1]{#1}

\usepackage{todonotes}

\newcommand\reals{\mathbb{R}}
\usepackage{algorithm}
\usepackage{algorithmic}
\usepackage{mathrsfs}

\usepackage{amsmath}
\usepackage{amsthm}
\usepackage{amsfonts}
\usepackage{amsmath}

\usepackage{xcolor}
\usepackage{graphicx}
\usepackage{cite}
\usepackage{soul}
\usepackage{subcaption}
\usepackage{todonotes}
\usepackage{hyperref}
\usepackage{mathtools}

\def\BibTeX{{\rm B\kern-.05em{\sc i\kern-.025em b}\kern-.08em
    T\kern-.1667em\lower.7ex\hbox{E}\kern-.125emX}}
\begin{document}

\title{
Automating Weak Label Generation for Data Programming with Clinicians in the Loop\\


{
\footnotesize }
}

\author{
\IEEEauthorblockN{
\revision{
Jean Park\IEEEauthorrefmark{1}\thanks{\IEEEauthorrefmark{1}Both authors contributed equally to this work.}, 
Sydney Pugh\IEEEauthorrefmark{1},
}\SP{TODO4}
Kaustubh Sridhar, 
Mengyu Liu, 
\revision{Navish Yarna}\SP{TODO2},\\
Ramneet Kaur, 
Souradeep Dutta, 
Elena Bernardis, 
Oleg Sokolsky, 
and Insup Lee
}
\IEEEauthorblockA{\textit{University of Pennsylvania} \\
Philadelphia, PA, USA \\
\{hlpark, sfpugh, duttaso, elber, lee\}@seas.upenn.edu}
}

\maketitle

\begin{abstract}

Deep learning with its unique ability to mine patterns in complex data has the ability to transform smart health. However, large Deep Neural Networks (DNNs) are often data hungry and need high-quality labeled data in copious amounts for learning to converge. This is a challenge in the field of medicine since high quality labeled data is often scarce. Data programming has been the ray of hope in this regard, since it allows us to label unlabeled data using multiple weak labeling functions. Such functions are often supplied by a domain expert. Data-programming can combine multiple weak labeling functions and suggest labels better than simple majority voting over the different functions. However, it is not straightforward to express such weak labeling functions, especially in high-dimensional settings such as images and time-series data. What we propose in this paper is a way to bypass this issue, using distance functions. In high dimensional spaces, it is easier to find meaningful distance metrics which can generalize across different labeling tasks. We propose an algorithm that queries an expert for labels of a few representative samples of the dataset. These samples are carefully chosen by the algorithm to capture the distribution of the dataset. The labels assigned by the expert on the representative subset induce a labeling on the full dataset, thereby generating weak labels to be used in the data programming pipeline. In our medical time series case study, labeling a subset of 50 to 130 out of 3,265 samples showed 17-28\% improvement in accuracy and 13-28\% improvement in F1 over the baseline using clinician-defined labeling functions. In our medical image case study, labeling a subset of about 50 to 120 images from 6,293 unlabeled medical images using our approach showed significant improvement over the baseline method, Snuba, with an increase of approximately 5-15\% in accuracy and 12-19\% in F1 score. 

\end{abstract}


\section{Introduction}

Deep learning has shown significant promise in various applications within the field of medicine, ranging from diagnostic imaging to drug discovery\cite{ching2018opportunities, piccialli2021survey, wang2019deep}. However, its widespread adoption and effectiveness is constrained by several challenges, especially the scarcity of labeled data\cite{nagendran2020artificial, willemink2020preparing}. Medical data, especially for rare conditions, is often limited\cite{rajpurkar2022ai}. Labeling the data requires expert-level knowledge, which can be time-consuming and expensive. Moreover, the rarity of certain medical conditions means that there are fewer instances or cases available for study, making it difficult to gather a sufficiently large dataset that deep learning algorithms require for effective training and validation. This scarcity of data can lead to sub-optimal models that do not adequately represent the full spectrum of the condition, potentially leading to less effective or even misleading outcomes when these models are applied in a clinical setting. Unfortunately, the assumption about the presence of abundant high quality labeled samples in medical settings, is often not the main concern when designing increasingly complex deep learning architectures. Thus it is imperative that the research community pays more attention to address this challenge.

To remedy the data scarcity challenge,  data programming has emerged as a potential alternative \cite{ratner2016data, karimi2020deep, ratner2020snorkel}. Data programming is a paradigm that aims to simplify and scale up the process of creating labeled training data, which is essential for supervised learning \cite{ratner2016data}. The dominant learning paradigm in medical AI, manual labeling of data, is often prohibitively expensive and time-consuming, especially for large datasets. Data programming offers a more cost-effective and scalable alternative\cite{dunnmon2020cross}. It allows domain experts to encode their knowledge in the form of \emph{rule of thumb} functions also known as weak labeling functions. Data programming uses these weak supervision signals to generate labels, which can be used for training a machine learning model, effectively bridging the gap between domain expertise and machine learning.

In reality, such weak labeling functions are heuristic or algorithmic rules that can assign labels to data points\cite{ratner2016data, ratner2017snorkel}. These functions are qualified as ``weak'' as a reminder to the fact that they are not as accurate or consistent as an expert labeling function can be. However, such heuristic rules are often present as domain knowledge and are amenable to being encoded easily. 
Developing weak labeling functions for data programming in machine learning involves a combination of domain expertise, data exploration, and heuristic approaches. Typically, experts provide valuable insights and identify specific patterns or indicators relevant to the labeling task. Alongside expert consultation, exploratory data analysis plays a vital role. It helps in uncovering hidden patterns, correlations, or frequent occurrences within the data, which can inform the creation of effective labeling functions.

However, it is challenging to construct weak labeling functions for high-dimensional data such as images and long horizon time series data. In order to be useful, the weak labeling function almost has to act like a decent classifier. It is well known that high-dimensional data samples contain vast amounts of information and variability. Therefore, it is hard to manually identify patterns and features across the dimensions. 
For instance, complicated features like texture may comprise subtle patterns that are not immediately obvious to the programmer. Even though experts can identify subtle textures, articulating this knowledge in a formal way that can be used to train machine learning models is challenging. The tacit knowledge that experts possess is not always easily transferable to weak-labeling functions.

To address this above challenge, we propose distance-based weak labels generation for high-dimensional data. Specifically, the contribution of this work are as follows:
\begin{enumerate}
    \item We show that using carefully chosen labeled samples to construct weak labels can significantly improve labeling accuracy. We demonstrate this with a thorough comparison against the weak label generation tool, \emph{Snuba}, for an equivalent number of labeled samples.
    \item We propose a novel algorithm which involves clinicians in the loop to label prototypical samples. We believe that our novel approach paves the way to clinician-AI collaboration in medical settings.
    \item We evaluate our algorithm on two challenging high-dimensional datasets -- (1) a time-series dataset consisting of 3,265 low \spotwo alarms collected as part of a study at a major hospital and (2) a dataset containing 6,293 de-identified medical images uploaded to a major U.S. hospital's electronic medical records labeled for specific body parts. The experimental results show the generated labeling function outperform the baselines by sufficient margins on both datasets.
\end{enumerate}
\section{Related Work}

\revision{
\subsection{Clinical Trials for Labeled Medical Datasets}
The gold standard for constructing large labeled medical 
datasets is a clinical trial, in which data is collected and manually labeled by clinical experts. 
Unfortunately, clinical trials require substantial time and effort, involving obtaining regulatory approvals, recruiting a diverse patient cohort, and labor-intensive and expensive manual labeling. 
For example, to create a dataset for training physiologic alarm suppression algorithms, nurses may need to review video feeds of patients to determine the actionability of alarms~\cite{macmurchy2017acceptability}. Such manual analysis can be extensive given the significantly high volume of physiologic monitoring alarms in hospitals, many of which are not actionable~\cite{paine2016systematic}. Our work is not meant to replace clinical trials; instead, we aim to reduce the amount of time spent manually labeling trial data by providing an semi-automated approach to labeling the data with the clinician in the loop.\SP{TODO15}
}

\subsection{Data Programming for Labeled Medical Datasets}

Recently, a novel method for efficiently and affordably annotating data has been introduced, termed data programming~\cite{ratner2016data}. This approach is especially relevant in the field of healthcare and is based on utilizing basic, quantitative insights to link data with its corresponding labels. For example, a healthcare professional might note that an alert should be highly prioritized if it concerns a patient over 65 years old with a heart rate exceeding 120 beats per minute for more than a minute. These observations are converted into algorithms known as \emph{labeling functions}, which may be imperfect, sporadically inaccurate, or even inconsistent. These functions either assign a specific class label to data or opt to `abstain' from labeling. By applying a diverse array of these labeling functions to an unlabeled dataset, data programming models generate labels for each data entry as probability distributions across the possible labels. The weak label for a data entry is then determined as the label with the highest probability, with confidence matching this probability.

The concept of data programming has gained significant attention as a solution for efficiently managing and labeling large datasets. An established technique in this area is Snorkel\cite{ratner2017snorkel, evensen2020data, denham2022witan}, which calculates the most effective weight for each labeling function through a generative graphical model, incorporating prior knowledge about the distribution of classes. This incorporation is crucial as it allows Snorkel to more accurately weigh the importance and relevance of each labeling function based on the known class distributions. By doing so, Snorkel effectively calibrates the labeling functions to better reflect the real-world scenarios they are intended to model, enhancing the accuracy and reliability of the labels generated for the dataset. This feature is particularly valuable in scenarios where class distributions are uneven or where some classes are significantly more prevalent than others, a common occurrence in many real-world datasets.

Labeling functions in weakly-supervised data labeling systems, such as Snorkel, are typically characterized by their inherent noise and variable error rates\cite{ratner2016data, ratner2019training}. This means that these functions, which are designed to assign labels to data points based on specific rules or heuristics, often produce labels that are not always accurate. The error rates can vary significantly between different labeling functions, depending on the complexity of the rules they apply and the quality of the underlying data they process. Moreover, these labeling functions may sometimes generate conflicting labels for the same data points\cite{varma2019learning}. This conflict arises because each labeling function operates independently, based on its own set of criteria or heuristics, without considering the outputs of other functions. For instance, one labeling function might label a data point as belonging to one class based on a particular feature or pattern, while another function might assign a different class to the same data point based on a different feature or pattern.

To address this issue, researchers explored the possibility of developing labeling generation methods that aggregate the noisy votes from multiple labeling functions\cite{ren2020denoising, lan2019learning, fu2020fast}. Researchers in \cite{ratner2016data, ratner2017snorkel, fu2020fast} propose a two-stage approach for weak labeling which using the training labels to train an end model for downstream tasks. Some researchers have attempted to develop a one-stage approach which integrates the label model with the end model\cite{lan2019learning, ren2020denoising}. It seeks to unify the process of labeling and model training, thereby reducing the complexity and potential errors associated with multi-stage processes.

Developing labeling functions for data programming can be challenging for domain experts, particularly for data such as time series and images. As a result, recent work have explored automated methods for generating labeling functions~\cite{varma2018snuba,das2020goggles,li2021weakly,zhao2021glara}. In the literature, the closest effort similar to our proposed approach is
\emph{Snuba}. It uses a small set of labeled examples (\ie a validation dataset) to automatically generate and refine such heuristic functions~\cite{varma2018snuba}. These functions are then applied to unlabeled data to create a training set. Snuba iteratively improves the functions based on their performance, thereby enhancing the quality of the labeled data over time. Snuba can help identifying the most informative features and patterns in the data and help creating new heuristic weak labeling functions which is feasible when there is limited access to domain expertise for writing labeling functions.

\section{Problem Formulation}

\begin{figure*}[th]
    \centering
    \includegraphics[width=\linewidth]{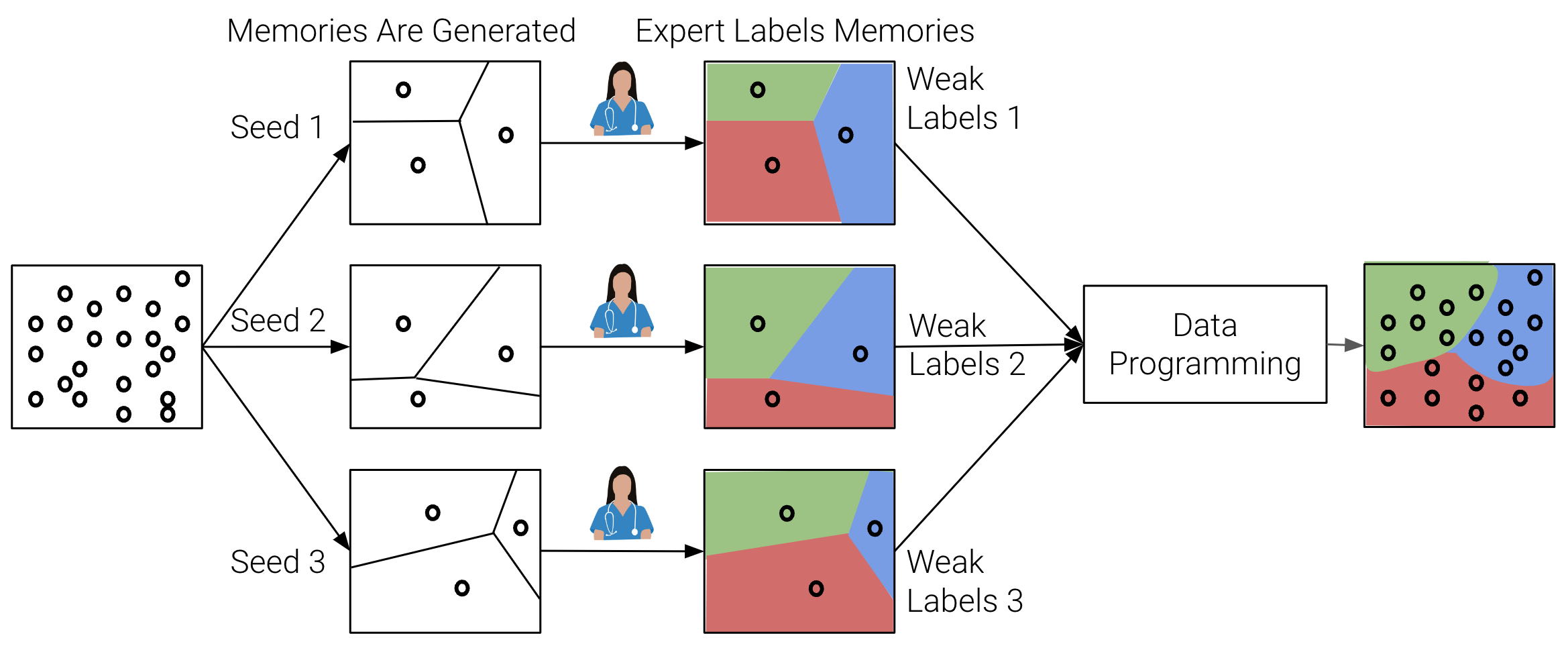}
    \caption{ Overall Approach: Starting from a dataset of unlabeled samples, we generate different partitions using different seeds. These partitions are centered around \emph{real} prototypical samples from the dataset referred to as \emph{memories}. Next, an expert clinician assigns a label to each prototype. This induces labels on the full dataset. Finally the  
    the memory-induced sets of weak labels are combined using a data-programming tool to arrive at better labels.
    }
    \label{fig:overview}
\end{figure*}


Assume that the inputs to be labeled belong to the space $\mathcal{X}$ and the possible labels are in the finite set $\mathcal{Y}$. We are given an unlabeled dataset $\mathcal{D}_u := \{ x_i \}_{i=1}^{N_u}$, such that each sample $x_i \in \mathcal{X}$ has a unique but unknown label in $\mathcal{Y}$. A vast majority of real world health data falls in this category. Note that in reality there exists an unknown oracle classifier $\mathcal{C} : \mathcal{X} \mapsto \mathcal{Y}$, which is capable of generating the ground truth labels $y_i = \mathcal{C}(x_i)$. However, $\mathcal{C}$ and therefore $y_i$ is unknown. 

Next, we discuss the concept of \emph{weak labels}. It is motivated by the potential use of quantitative intuitions about how the data corresponds to labels. To restate a familiar example, a clinician might say, ``when a patient is over 70 years old and has had a heart rate over 125 beats for over a minute such an alarm is pretty likely of being high priority''. 
A weak label is defined as $\hat{y}_i = \mathcal{C}^w(x_i)$, where $\mathcal{C}^w$ is a weak labeling function.
Such weak labels are  a source of \emph{weak supervision} to the learning algorithm. One often gains from generating multiple such weak labels $\{ \hat{y}_i^1, \hat{y}_i^2, \dots, \hat{y}_i^{N_w} \}$ corresponding to each input sample $x_i$, assuming that there are multiple experts ($N_w$) who are willing to suggest such weak labels. The next step is to automatically combine such weak labels and therefore help achieve better labels for an unlabeled sample $x_i$. This is known as  data-programming \cite{ratner2016data} in the literature.  Such weak labels sourced from multiple experts can be combined by tools such as \emph{Snorkel} to suggest a single probability distribution and therefore a single possible label corresponding to each unlabeled sample. 

Data programming typically advocates for the use of weak-labeling functions $\mathcal{C}^w$, which are produced by an expert, to automatically generate a weak label $\hat{y}_i^j$ for an input sample $x_i$. However, the challenge in recent years has been to come up with such weak labeling functions. Especially for high dimensional data like long-horizon time series signals of ICU patient vitals or diagnostic images for different pathological conditions. For instance, a weak labeling function for an image of a skin lesion would mean coming up with an algorithm which can nearly classify images to different categories in $\mathcal{Y}$. This is generally considered a hard problem, especially in the absence of large quantities of labeled samples. But, addressing the weak label generation problem can enable us to use the remaining data-programming pipeline, and build better machine learning models for smart health.

Before we state our problem, we make the following two assumptions: $1.$ We have access to a distance function $d : \mathcal{X} \times \mathcal{X} \mapsto \reals $, which can capture the degree of similarity between input pairs; and $2.$ An expert has the ability to label a small subset ($< N_L$) of samples with the ground truth.  We can now state our research question : 

\emph{Given a set of unlabeled samples $\mathcal{D}_u$, a distance function $d$, and an ability to label at most $N_L$ samples, is it possible to automatically generate $N_w = \big\lfloor \frac{N_L}{|\mathcal{Y}|} \big\rfloor $ sets of weak labels for the unlabeled samples in $\mathcal{D}_u$? }\\

We will answer this question in detail over the next few sections. In Section \ref{sec:overall_approach} we give a brief overview of the steps involved at a high level. In Sections \ref{sec:distance-for-time-series} and \ref{sec:distance-for-image} we go over the details of the different distance functions we used for two different data types in medicine. Finally, in Section \ref{sec:results}, we demonstrate the effectiveness of our method. 

\section{Overall Approach} 
\label{sec:overall_approach}





We outline our overall approach in this section as shown in Figure \ref{fig:overview}. To restate, we begin with an unlabelled set of samples $\mathcal{D}_u$ and a distance function $d$ which captures the degree of similarity. Additionally, we have an expert which can give us access to ground truth up to a budget of $N_L$ samples.  The main essence of our algorithm is that we carefully figure out \emph{which} data-samples need to be labeled in order to induce a labeling on the remaining samples. We outline the overall algorithm next: 

\begin{enumerate}
    \item Learn representative samples called memories from $\mathcal{D}_u$, with seed $s^k$. Let this set of memories be $M^k \subset \mathcal{D}_u$, such that $M^k := \{ m_1, m_2, \dots, m_{\lvert M^k \rvert} \}$.
    
    \item This partitions the unlabeled dataset into $\lvert M^k \rvert$ different groups $\{ g_1, \dots, g_q, \dots, g_{\lvert M^k \rvert} \}$ such that $\mathcal{D}_u = \overset{\lvert M_k \rvert}{\underset{q = 1}{\bigcup}} g_q$ according to a nearest neighbor sense of partitioning, that is, $x_i \in g_q$ iff $q = \underset{1 \leq k \leq \lvert M^k \rvert }  {\arg\min} \; d(x_i, m_k)$. This is termed as the \emph{Memory Generation} step in Figure \ref{fig:overview}.
    
    \item Query a clinician to label all samples in $M_k$. That is, $\forall m_q \in M_k$, we obtain a $y_q = C(m_q)$. This induces a weak label on all the elements in group $g_q$. Essentially, $\forall x_j \in g_q$ we assign a weak label $ \revision{\hat{y}_j^{(k)}} = C^w(x_j) = y_q$. Drawing intuitions from Figure \ref{fig:overview}, choosing a color (labels) on the memory points induces a color on all the samples in a partition.

    \item Repeat steps $1$ to $3$ for $N_w$ different seeds $s_1, \dots, s_k, \dots s_{N_w}$.

    \item The weak labels according to the different seeds are inputs to the data-programming tool which creates the resulting labels. \revision{Each seed $s_k$ acts as a distinct labeling function, producing outputs $\{\hat{y}_j^{(k)}\}$. Data programming combines the outputs of these $N_w$ labeling functions for each sample $\left\{\hat{y}_j^{(1)}, \dots, \hat{y}_j^{(k)}, \dots, \hat{y}_j^{(N_w)}\right\}$ into a single probabilistic label $[p_1, \dots, p_{|\mathcal{Y}|}]$ where ${\sum_{i=1}^{|\mathcal{Y}|} p_i = 1}$.}\SP{TODO5}

\end{enumerate}

We discuss the details of the memory generation algorithm in Section \ref{sec:memory-generation}. However, a pertinent detail to note here is that the number of memories picked in Step $1$ in the set $M^k$ depends on a hyper-parameter called the distance threshold $t$. This serves as a parameter which indirectly controls the number of memory groups we end up with. We pick a threshold $t$ such that, there are at least $|\mathcal{Y}|$ memories in $M^k$, for all the seeds $k$. That is,  $$\lvert M^k \rvert \geq |\mathcal{Y}|.$$
This is because if $\lvert M^k \rvert < |\mathcal{Y}|$, then it is guaranteed that there is at least one sample in $\mathcal{D}_u$ which is wrongly labeled. Assuming $\mathcal{D}_u$ has at least one sample from each class. Let the total number of samples to be labeled by the expert be $N_s$. Then, 
$$N_s = \underset{1 \leq k \leq N_w}{\sum} \lvert M^k \rvert.$$
Therefore, the following serves as a lower bound on $N_s$:
$$N_s \geq N_w \times |\mathcal{Y}|.$$
Since, the maximum labeling budget is $N_L$, then,
$$N_L \geq N_s \geq N_w \times |\mathcal{Y}|.$$

Therefore the number of weak labels is $N_w \leq \big \lfloor \frac{N_L}{|\mathcal{Y}|} \big \rfloor $. Thus, in the worst case (performance wise) the expert is expected to  label at least $|\mathcal{Y}|$ samples and the lower bound on $N_L$ grows in proportion to the number of weak label sets.


\section{Methodology}
\subsection{Memory Generation Algorithm}
\label{sec:memory-generation}

Unsupervised clustering, from simple k-means like algorithms to self-organizing maps (SOM) and complex neural gas algorithms \cite{hartigan1979algorithm, kohonen1990self, martinetz1991neural, mcnn, sridhar2023guaranteed_conformance}, is a promising way to capture a distribution of the dataset.  In a fashion similar to \textit{k-means} clustering, we wish to form partitions of the data into distinct groups or clusters.  Clustering with \textit{k-means} is a well-known tool but has its challenges when used in the context of images. An issue with \textit{k-means} is that it can potentially produce virtual cluster centers which are absent in the original dataset. This is essentially because a simple mean of two (or more) data points (for instance images in our case)  might not correspond to a real data point. This is crucial since we wish to use these centers and query an expert for the labels. A sure shot to ensure \emph{realistic} data samples is to pick from the dataset itself. Another potential issue with vanilla k-means is that it is often susceptible to outliers in the data. Hence, we restrict ourselves to partitioning around points from the unlabeled set $\mathcal{D}_u$. \\

The  closest algorithm which achieves this is PAM \cite{pam}, which is short for \emph{Partitioning Around Medoids}. Intuitively the algorithm tries to search for centrally located data samples called \emph{medoids} which are used to define the cluster boundaries in a nearest medoid sense. For the distance function $d: (x_1,x_2) \mapsto \reals$ on the data set $\mathcal{D}_u$, PAM tries to select a set of $r$ medoids - $ M_r: \{m_1, m_2, \dots , m_r \}$ such that the following cost is minimized:
\begin{equation}
\label{eq:PAM_cost}
Cost(M_r) = \sum_{i=1}^{n} \underset{m_j \in M_r}{min} d(m_j, x_i).
\end{equation}
We assume that the inner minimization is always possible, and we are able to break ties arbitrarily among distinct members of the set $\mathcal{S}$. The challenge with PAM is that the naive implementation has a runtime complexity of $O(N_u^2r^2)$ \cite{pam_compare}. Even though there exists faster variants, they are still largely inaccessible for applications at the scale of image data sets generated in medical settings. 

In order to circumvent this challenge we use a variant of the \emph{Clustering Large Applications based upon Randomized Search} (CLARANS) \cite{clarans} algorithm introduced in \cite{memory_paper, memory-statistical-guarantee-paper}. For reference we outline this process in Algorithm \ref{alg:learn_memories},
which combines randomized global search with a local clustering method. The algorithm aims to minimize the objective in Equation \ref{eq:PAM_cost} and returns a subset of the training points. These are referred to as \emph{memories} from here on. We discuss the working of this algorithm in further detail next. \\

\begin{algorithm}[t]
\caption{Generate Memories}
\label{alg:learn_memories}
\flushleft
\textbf{Input: } $\mathcal{D}_u$ : $ \{x_1, x_2, \dots x_{N_u}\} $  \\
\textbf{Output: } Memories $ \mathcal{M} : \{ m_1, m_2, m_3, \dots, m_r \}  $ \\
\textbf{Parameter} : (Max Global Steps : $Z_g$ , Max Local Steps : $Z_l$, Distance Threshold $t$, Random Seed $s$)
\begin{algorithmic}[1]
\STATE BestCost = $\infty$
\FOR{$ 1 \leq g \leq Z_g$}
\STATE Memory Set $M$ = GenerateInitialMemories($\mathcal{D}_u,t,s,g$)
\STATE CurrentCost = ComputeCost($M$)
\FOR{$1 \leq l \leq Z_l $}
\STATE Perturb the memories in $M$ locally to generate $M^\prime$  
\STATE NewCost $\leftarrow$ ComputeCost($M^\prime$)
\IF{NewCost $<$ CurrentCost }
\STATE $M \leftarrow M'$
\STATE CurrentCost $\leftarrow$ NewCost
\ENDIF
\ENDFOR
\IF{CurrentCost $<$ BestCost}
\STATE BestCost = CurrentCost
\STATE $\mathcal{M}$ = $M$
\COMMENT{Store the best memory set observed}
\ENDIF
\ENDFOR
\RETURN $\mathcal{M}$
\end{algorithmic}
\end{algorithm}

\noindent \textbf{Algorithm to Generate Memories: }The algorithm starts with a reasonable choice for initial memories $M$ using $GenerateInitialMemories$. The aim of this function is to ensure that each sample in $\mathcal{D}_u$ is within a distance of at most $t$ from a memory in  $M$. These memories are randomly chosen according to the seed $s$, but ensure coverage of all points. Notice that we do not choose the number of memories a priori but instead gets picked as a consequence of \textit{distance score}  ${d}$. Next, starting from this initial choice, the inner loop (Lines $5-12$) greedily looks for local improvements for a fixed number of iterations $Z_l$. The local improvements involve switching a memory for a nearby data sample as a candidate replacement. The partitioning cost for the choice of memories is computed by the function $ComputeCost$ which evaluates Equation \ref{eq:PAM_cost}. The outer loop of the algorithm (Lines $2-17$) resets the initial choice repeatedly for different iterations (Line $3$) and keeps track of the memory set $\mathcal{M}$ with the minimum cost for each such reset. This set $\mathcal{M}$ is returned in the end. Algorithm  \ref{alg:learn_memories} trivially terminates since each search proceeds for a fixed number of steps.\\

Thus, depending on the type of data, Algorithm \ref{alg:learn_memories} should use different distance functions $d$ to choose the relevant memories. This is what we discuss next, where depending on the domain we choose the appropriate distance function. This is discussed in detail for  time-series and image data in Section \ref{sec:distance-for-time-series} and \ref{sec:distance-for-image}, respectively.

\subsection{Distance Metric for Medical Time Series Data} 
\label{sec:distance-for-time-series}



In this case study, our objective is to label physiologic monitoring alarms with respect to their suppressibility. Physiologic monitors that continuously measure parameters like blood oxygen, heart rate, and respiratory rate often overwhelm clinicians with a large amount of false alarms causing alarm fatigue~\cite{paine2016systematic}. Ideally, clinicians should only be alerted by the alarms  that provide informative or actionable insights (\ie \emph{non-suppressible} alarms), while the rest of the alarms are silenced (\ie \emph{suppressible} alarms). Achieving this requires an alarm suppression system capable of determining the suppressibility of a potential alarm based on the patient's vitals (time series) at the time of the alarm. However, the system must be trained and evaluated prior to deployment which requires labeled time series alarm data. Traditionally, acquiring such data involves an observational study followed by manual labeling, which is expensive and time-consuming. Hence we aim to apply data programming to overcome this challenge. Unfortunately, defining labeling functions for medical time series data is generally challenging.


Formulating quantitative labeling functions can be especially challenging for clinicians, given the dynamic and intricate nature of physiological data. For instance, a clinician might describe the vitals time series as ``jagged'' or ``erratic'', which are more of a qualitative observation than a quantifiable metric. 
Moreover, medical time series data is inherently dynamic and can exhibit complex patterns. 
Designing labeling functions that capture the nuances of these patterns and their clinical significance can be a complex task. One common approach involves extracting features from a time series and defining labeling functions over them, but these primitives are not often well-defined. In contrast, devising a distance metric provides a more tractable path, leveraging mathematical principles to objectively quantify relationships between data points.


We leverage the Dynamic Time Warping (DTW) distance~\cite{berndt1994using} to assess the similarity between medical time series data. DTW proves to be a robust choice for comparing medical time-series data due to its ability to handle variable speeds and temporal misalignment inherent in physiological signals. Unlike traditional metrics that assume a fixed alignment between sequences, DTW allows for flexible time warping, making it well-suited for capturing the complex dynamics of vital sign data where patients may exhibit variations in the timing of physiological events.

\subsection{Distance Metric for Medical Image Data}
\label{sec:distance-for-image}

\begin{figure}[t]
  \centering
  \begin{subfigure}{0.23\textwidth}
    \centering
    \includegraphics[width=\textwidth]{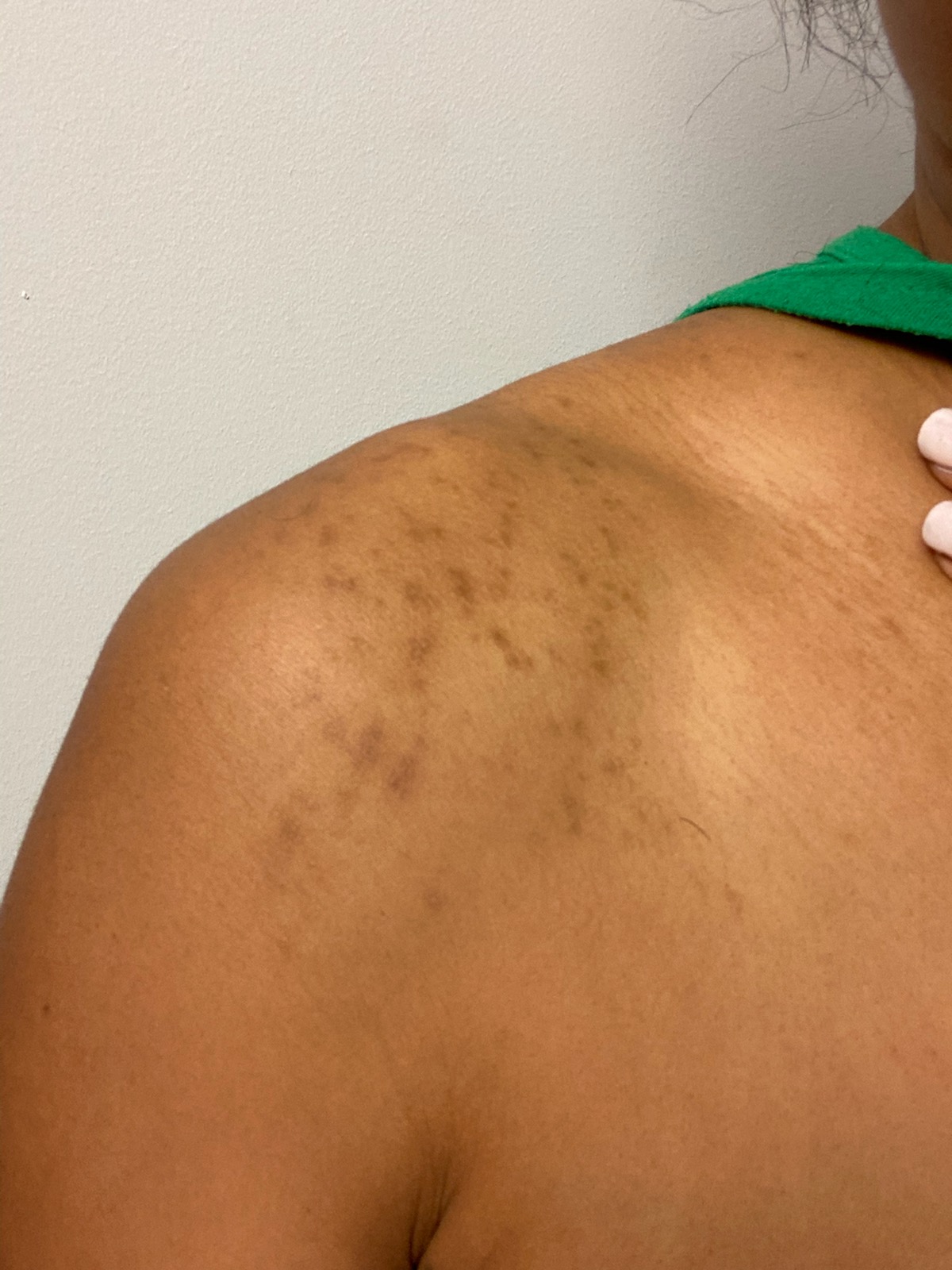}
    \caption{Neck and/or Shoulders}
    \label{fig:sub1}
  \end{subfigure}%
  \hfill
  \begin{subfigure}{0.23\textwidth}
    \centering
    \includegraphics[width=\textwidth]{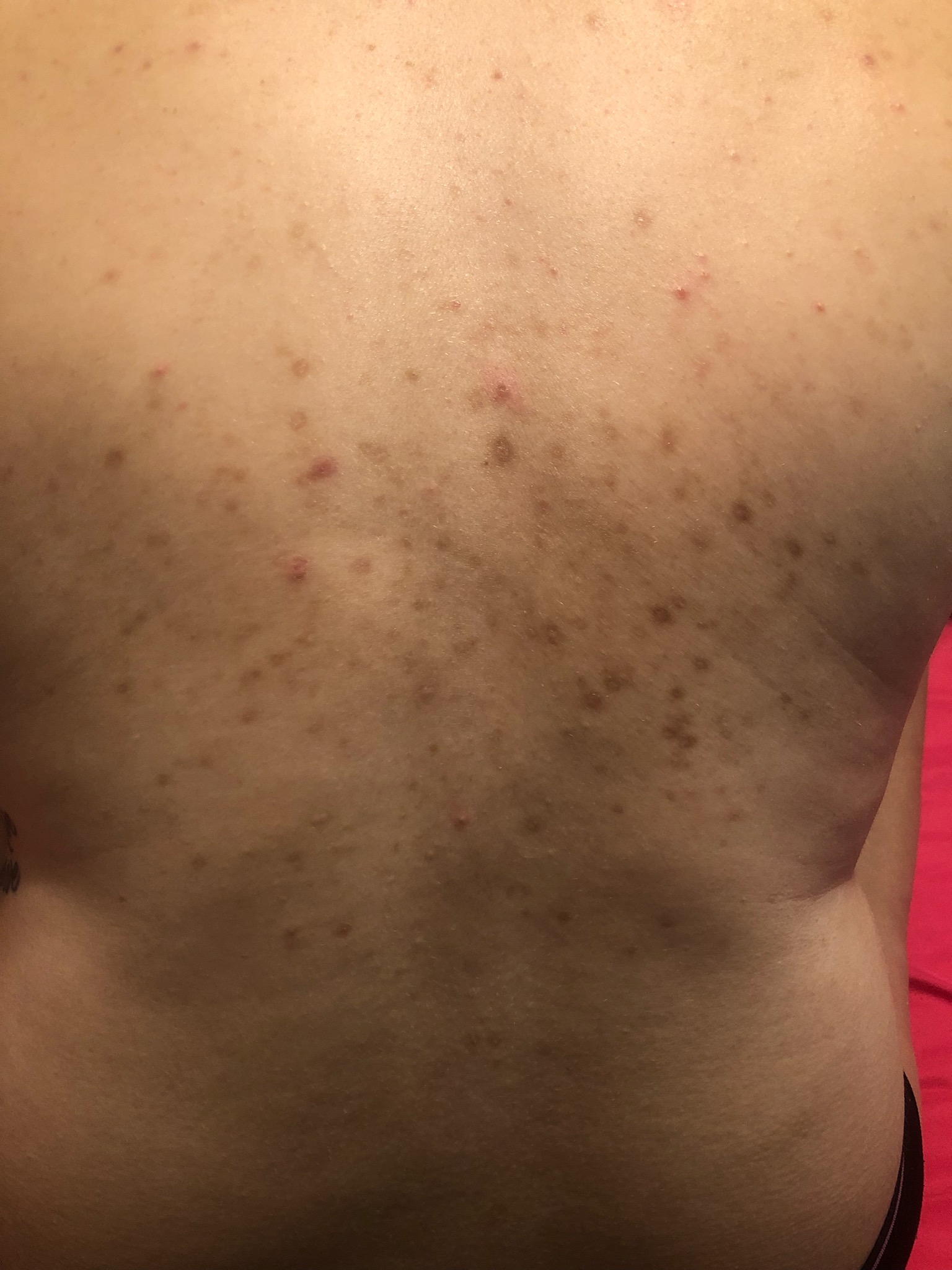}
    \caption{Back}
    \label{fig:sub2}
  \end{subfigure}%
  \hfill
  \begin{subfigure}{0.23\textwidth}
    \centering
    \includegraphics[width=\textwidth]{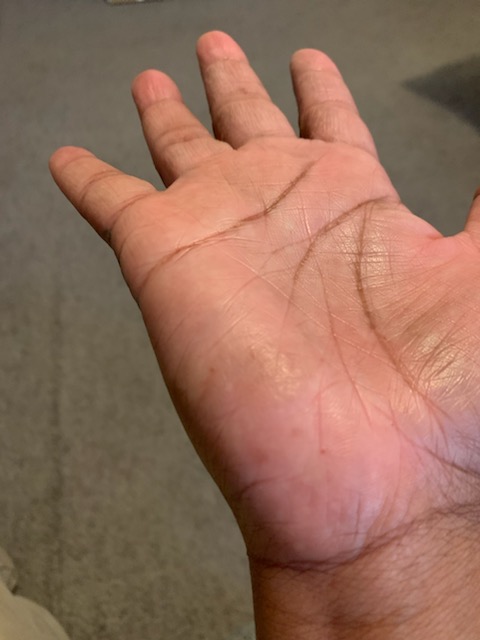}
    \caption{Hand}
    \label{fig:sub3}
  \end{subfigure}%
  \hfill
  \begin{subfigure}{0.23\textwidth}
    \centering
    \includegraphics[width=\textwidth]{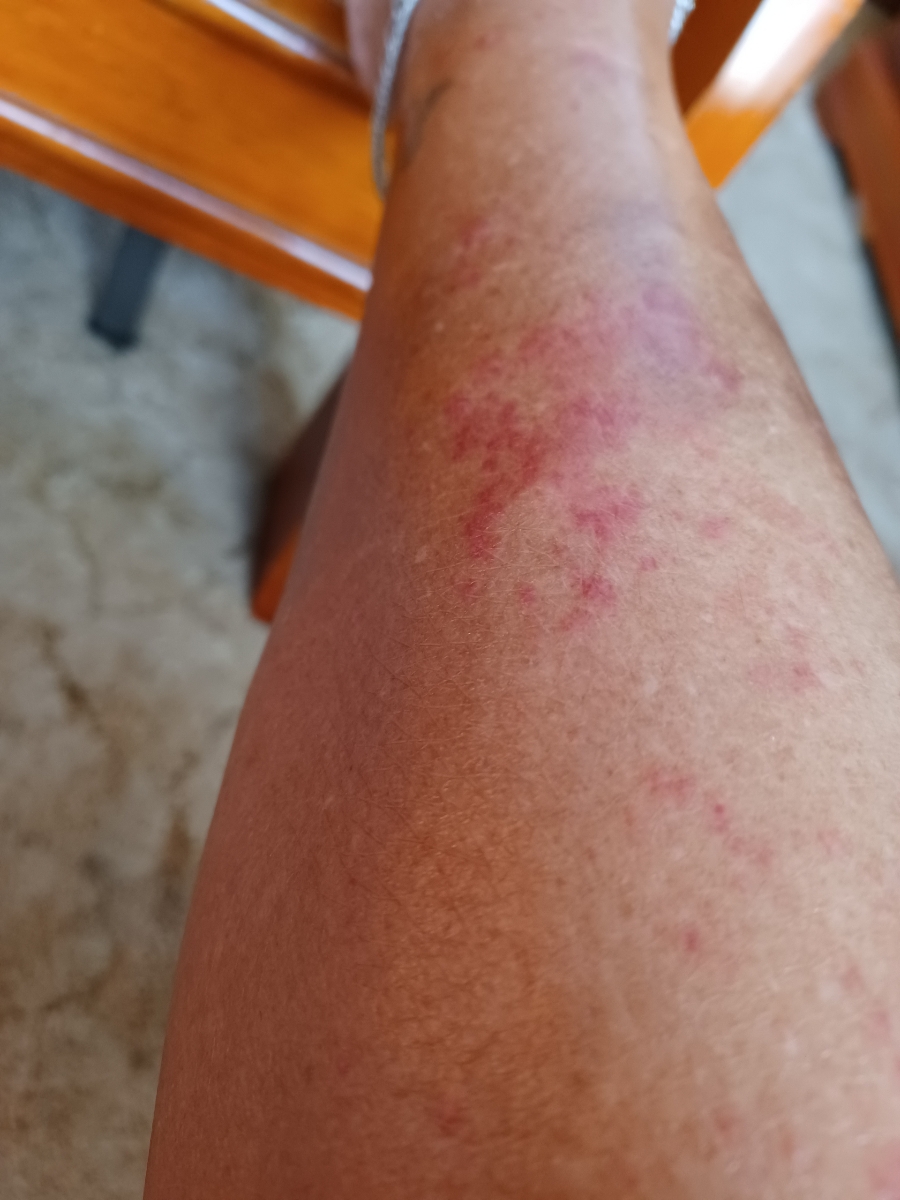}
    \caption{Leg}
    \label{fig:sub4}
  \end{subfigure}
  \caption{Medical image data examples. Sample dermatological images taken by patients for teledermatology consultation.}
  \label{fig:four_images}
\end{figure}



\revision{
In this case study, we address the challenge of identifying specific body parts in dermatological images captured by patients via smartphone cameras. Advances in smartphone camera quality have played a significant role in improving the reliability in the diagnosis of dermatological conditions through teledermatology~\cite{barbieri2014reliability}. 
There is considerable interest in teledermatology to develop algorithms to assist with the clinical workflow and improve efficiency. However, developing such algorithms generally requires a large and diverse annotated dataset for training, which can be time-consuming and costly to obtain due to the need for expert annotations.
For example, the location of lesions on the body helps narrow a diagnosis, as many skin conditions are specific to certain body areas. 
Annotating body parts for a large dataset of dermatological images, however, can be tedious and challenging, particularly when images are captured by patients in uncontrolled environments (\eg home, car, etc.)~\cite{sitaru2023automatic}.
Data programming is a promising approach for overcoming these challenges associated with manual annotation of body parts in dermatological images, but defining labeling functions for medical image data also presents its own challenges.
}\SP{TODO1}



In the context of high-dimensional data like images, defining and applying labeling functions that capture human visual understanding across different kinds of images is very challenging. In contrast, distance metrics provide a straightforward way to quantify similarity between images based on their feature representations and can be easily scaled to large image datasets. Distance functions also reduce human subjectivity in the labeling process, leading to a more consistent and objective analysis. \revision{Such distance functions can be leveraged by our approach to automatically generate labeling functions for image data, consequently facilitating an easier application of data programming.}

Here, we employed the Contrastive Language–Image Pretraining (CLIP) \cite{radford2021learning} model to extract two types of features for our analysis: image representations and probability distributions across image labels. CLIP, a foundation model trained on an internet-scale dataset of images and text, is able to generalize to diverse datasets and effectively identify image-text similarities. We harness CLIP to generate the probability distribution for each image in our dataset against the 10 predefined body part labels. In addition, we used CLIP's image encoder to extract intermediate image representations with a vector length of 512.


To evaluate the similarity of the predicted probability distributions over labels, we applied the Kullback-Leibler (KL) divergence, a method for measuring the distance between two probability distributions.  For the image representations, we used the Euclidean distance metric to assess the visual similarity between images.

\section{Results }
\label{sec:results}

In this section, we evaluate our approach on medical time series and medical image case studies. 

\subsection{Implementation}

For data programming, we use the implementation of Snorkel and majority vote provided in the Snorkel Python library, version 0.9.7 (www.snorkel.org). To calculate DTW distance, we use the Python library \texttt{DTAIDistance}~\cite{meert2020dtaidistance}.

\subsection{Results for Medical Time Series Data}

\subsubsection{Dataset}

We evaluate our approach on a low oxygen saturation (\ie \spotwo low) alarm dataset extracted from a deidentified dataset originally collected as part of a study approved by the Institutional Review Board of a hospital.
Researchers video-recorded 551 hours of patient care on a medical unit at 
a major US hospital 
during July 2014 to November 2015 from 100 children whose families and nurses provided consent. In addition, the following data was collected: patient background information, all physiologic monitoring alarms with corresponding timestamps, and continuously recorded vital signs. 

To obtain the dataset of \spotwo low alarms, we first identify the maximum length of the \spotwo low alarms. For each alarm, we then extract the \spotwo time series within a window of that determined length, positioning the alarm in the middle of the window. In total, 3,265 \spotwo low alarms were extracted as part of this dataset.

Each alarm is annotated in terms of technical/clinical, valid/invalid, and actionable/non-actionable alarms [3]. We interpret these annotations with respect to suppressibility as follows: technical, valid non-actionable, and invalid alarms are interpreted as suppressible, whereas only valid actionable alarms are non-suppressible.


\subsubsection{Baseline}

We compare labels output by our approach to labels obtained by applying data programming with weak labels produced by domain-expert defined labeling functions. We use the set of sixty-two clinician-designed labeling functions developed for this example from \cite{pugh2022evaluating, pugh2021high}. The labeling functions analyze the time series data to make predictions on suppressibility, \textit{e.g.,} an alarm is non-suppressible if the heart rate is above 220 for longer than 10 seconds after the alarm starts, otherwise it abstains.

\subsubsection{Result}

\begin{table*}[t]
    \centering
    \begin{tabular}{lcccccc}
        \toprule
        \multirow{2}{*}{Metric} & \multicolumn{2}{c}{Clinician-} & \multicolumn{4}{c}{Our Approach} \\
        & \multicolumn{2}{c}{Defined LFs} & \multicolumn{2}{c}{$N_L=54$} & \multicolumn{2}{c}{$N_L=134$} \\
        & Majority Vote & Snorkel & Majority Vote & Snorkel & Majority Vote & Snorkel \\
        \midrule
        Accuracy & 0.630 & 0.467 & \underline{0.807} & 0.746 & \textbf{0.808} & 0.642 \\
        F1 & 0.734 & 0.536 & \underline{0.872} & 0.823 & \textbf{0.873} & 0.728 \\
        \bottomrule
    \end{tabular}
    \caption{Performance of labels generated by our approach with varying labeling ability $N_L$ compared to labels from data programming with weak labels from clinician-defined labeling functions. We highlight the best performance scores across approaches in bold, and underline the second best scores.}
    \label{tab:alarms}
\end{table*}
 
Table~\ref{tab:alarms} shows the accuracy and F1 of labels produced by our approach with the number of labeled samples ($N_L$) of 54 and 134 samples compared to that of the baseline using clinician-defined labeling functions. \revision{To produce these results, we set the max global steps $Z_g$ to 5 and the max local steps $Z_l$ to 30, and the distance threshold $t$ to 90 and 60 to yield $N_L = 54$ and $N_L = 134$ samples to be labeled by the clinician, respectively.}\SP{TODO12} The results demonstrate that our approach yields greater accuracy and F1 scores than the baseline approach. When using majority vote we observe an improvement of approximately 17\% in accuracy and 13\% in F1 using our approach. Using Snorkel, we observe improvement of up to 28\% on both accuracy and F1 using our approach. The results in Table~\ref{tab:alarms} demonstrate that leveraging a small number of labeled examples supplied by a clinician can yield more accurate labels.

\subsection{Results for Medical Image Data}

\subsubsection{Dataset}
The dataset used in our study consists of 10,140 de-identified images, retrospectively collected from a teledermatology database. These images were captured by patients using mobile devices and uploaded into their electronic medical records at a major US hospital. For efficient labeling, we utilized the LabelMe\cite{russell2008label} interface, adapted to identify specific body parts. Labels were based on clinical interest, covering a range of body parts such as the `Face', `Scalp', `Ears', `Torso', `Torso-Chest', `Torso-Abdomen', `Breasts', `Back', `Neck \& Shoulders', `Arms', `Armpits', `Hands', `Legs', `Feet', `Genitalia', `Buttocks', `Skin-only', and `Not Derm' for images that were not related to dermatology. This process was focused on regions of interest (ROI) within each image.

The labeling was conducted by two annotators, with any discrepancies resolved through consensus or by marking the image as unclear, ensuring no ambiguities in body part identification remained. During the labeling phase, images with multiple ROIs had all recognizable body parts labeled. However, due to the complexities associated with multi-label classification in our label generation algorithm, we later excluded these multi-labeled images from the dataset. We also encountered images with no labels, either due to unrecognizable locations or the presence of rare body parts not defined by the labels, such as the inside of the mouth. These images were also dropped from the dataset. In addition, we merged `Torso-Abdomen' and `Torso-Chest' images into a single `Torso' label. We removed `Skin-only' and `Not derm' labels as they were not related to body detection. Furthermore, `Ears', `Breasts', `Genitalia', and `Buttocks' were discarded due to their low frequencies, each having less than 100 images. After refining the dataset, 6,293 images with single labels were retained for analysis. The distribution of these images across the various body parts is shown in Table~\ref{tab:image_dataset_distribution}.

\begin{table}[t]
    \centering
    \begin{tabular}{lc}
        \toprule
        \multirow{2}{*}{Classes} & Number of   \\
        & Images   \\
        \midrule
        Face & 937  \\
        Scalp & 998 \\
        Torso & 299 \\
        Back & 381  \\
        Neck and/or Shoulders & 352  \\
        Arms & 702  \\
        Armpits & 185 \\
        Hands & 1,066 \\
        Legs & 901  \\
        Feet & 472  \\
        \midrule
        Total & 6,293  \\
        \bottomrule
    \end{tabular}
    \caption{Medical image dataset class distribution.}
    \label{tab:image_dataset_distribution}
\end{table}

\subsubsection{Baseline}
We use Snuba~\cite{Varma2018SnubaAW} as a baseline method for our experiment. Snuba is a system designed to overcome the challenges of gathering high-quality training labels for deep learning models. Snuba automatically generates the heuristics that labels a small subset of the data for which it is most accurate and applies this process iteratively to a large unlabeled dataset. 

\subsubsection{Result}

\revision{We transform our multi-class dataset into ten one-versus-all labeling problems and apply our approach to each class individually. This setup enables a direct comparison with the Snuba baseline, as Snuba's implementation currently only supports a binary setting. We report the accuracy and F1 of the labels produced by our approach using CLIP image representations and probability distributions for varying labeling budgets $N_L$ in Table~\ref{tab:images_class}. To produce the results for image representations, we set the max global steps $Z_g$ to 5 and the max local steps $Z_l$ to 30, and the distance threshold $t$ to 0.875 and 0.825 to yield $N_L = 59$ and $N_L = 113$ samples to be labeled by the clinician, respectively. For probability distributions, we set the max global steps $Z_g$ to 5 and the max local steps $Z_l$ to 30, and the distance threshold $t$ to 0.8 and 0.5 to yield $N_L = 54$ and $N_L = 116$, respectively. We summarize our results by computing the average accuracy and average F1 score across all ten one-versus-all labeling problems, which are presented in Table~\ref{tab:images1} and Table~\ref{tab:images2}. The results show our approach, using both majority vote and Snorkel, generally demonstrates improvement over Snuba. Using image representation features, our approach yields improved accuracy over Snuba (up to 6\%). Unfortunately, we do not observe consistent improvement in F1 score over Snuba, but the F1 scores are generally comparable. In contrast, for probability distribution features, our approach consistently outperforms Snuba in both accuracy and F1. We observe a significant increase in accuracy by 5-15\% and in F1 score by 12-19\%.}

Moreover, our results show that our approach using majority vote outperforms our approach using Snorkel in terms of both accuracy and F1 scores.\SP{@Jean Can you add some intuition why.} \revision{Given that about half of the 10 classes have a conflict rate above 10\% among labeling functions, majority voting tends to align with the correct label more often than Snorkel's probabilistic model, despite these disagreements.} Furthermore, when analyzing the features, the probability distribution feature using KL divergence consistently outperforms the Euclidean distance metric on image representation features.\SP{@Jean Again, try to add some intuition why (after the next sentence).} This is evident even when the number of labeled samples ($N_L$), determined by adjusting the distance threshold, 
are roughly the same as those used for image representation. \revision{We believe KL divergence effectively captures the subtle nuances in class probability distributions, whereas Euclidean distance measures the absolute differences in feature values, potentially overlooking non-linear perceptual differences.}

\revision{Finally, we apply our approach to perform multi-class labeling of the dataset. Table~\ref{tab:multiclass_performance} shows the F1 scores of the labels produced by our approach compared to that of the top-1 predicted label output by CLIP.}
CLIP's performance is inconsistent, either excellent or extremely poor, achieving an overall weighted F1 score of 52.7\% and an overall accuracy of 55.2\%. On the other hand, our method shows more balanced distribution across classes. It performs better on classes where CLIP struggles, such as Back, Neck and/or Shoulders, and Armpits. Our approach generally yields higher overall F1 score and accuracy, with the exception of when we use majority vote with CLIP image representations and labeling budget $N_L=59$. In this case the overall accuracy is marginally lower than CLIP's. However, our method using CLIP's probability distribution for the majority vote with $N_L=54$ shows the best results, with a 12.1\% higher accuracy and a 14.5\% higher F1 score compared to CLIP.


    


\begin{table}[t]
    \centering
    \begin{tabular}{llccc}
        \toprule
        \multirow{2}{*}{$N_L$} & \multirow{2}{*}{Metric} & \multirow{2}{*}{Snuba} & \multicolumn{2}{c}{Our Approach} \\
        & & & Majority Vote & Snorkel \\
        \midrule
        59 & Accuracy & 0.883 (0.019) & \textbf{0.930} (0.032) & \underline{0.898} (0.066) \\
        & F1 & \textbf{0.431} (0.031) & \underline{0.410} (0.288) & 0.401 (0.268) \\
        \midrule
        113 & Accuracy & 0.869 (0.009) & \textbf{0.931} (0.030) & \underline{0.898} (0.048) \\
        & F1 & \underline{0.454} (0.010) & \textbf{0.459} (0.263) & 0.426 (0.246) \\ 
        \bottomrule
    \end{tabular}
    \caption{Performance comparison between weak labels generated by Snuba and our approach using \textbf{Euclidean distance} on CLIP image representation}
    \label{tab:images1}
\end{table}

\begin{table}[t]
    \centering
    \begin{tabular}{llccc}
        \toprule
        \multirow{2}{*}{$N_L$} & \multirow{2}{*}{Metric} & \multirow{2}{*}{Snuba} & \multicolumn{2}{c}{Our Approach} \\
        & & & Majority Vote & Snorkel \\
        \midrule
        54& Accuracy & 0.788 (0.004) & \textbf{0.934} (0.038)& \underline{0.918} (0.058)\\
        & F1 & 0.320 (0.005) & \textbf{0.515} (0.335) & \underline{0.492} (0.359)\\
        \midrule
        116& Accuracy & 0.891 (0.004) & \textbf{0.941} (0.027)& \underline{0.916} (0.046)\\
        & F1 & 0.453 (0.007) & \textbf{0.579} (0.264)& \underline{0.566} (0.288)\\
        \bottomrule
    \end{tabular}
    \caption{Performance comparison between weak labels generated by Snuba and our approach using \textbf{KL divergence} on CLIP probability distribution feature}
    \label{tab:images2}
\end{table}

\begin{table*}[t]
    \centering
    \begin{tabular}{l|l|cc|cc|cc|cc}
        \toprule
        \multirow{4}{*}{Positive Class} & \multirow{4}{*}{Metric} &  \multicolumn{8}{c}{Our Approach} \\
         & &  \multicolumn{4}{|c|}{CLIP Image Representation} &  \multicolumn{4}{c}{CLIP Probability Distribution} \\
        &  & \multicolumn{2}{|c|}{$N_L=59$} & \multicolumn{2}{|c|}{$N_L=113$} & \multicolumn{2}{|c|}{$N_L=54$} & \multicolumn{2}{c}{$N_L=116$} \\
        & & Majority Vote & Snorkel & Majority Vote & Snorkel & Majority Vote & Snorkel & Majority Vote & Snorkel\\
        \midrule
        Face & Accuracy  & 0.921& 0.786 & 0.928& 0.830 & 0.897 & 0.815 & 0.903 & 0.924\\
        & F1  & 0.650 & 0.560 & 0.723& 0.608 & 0.566 & 0.562 & 0.535 & 0.724\\
        Scalp & Accuracy  & 0.954 & 0.898 & 0.95& 0.918 & 0.962 & 0.961 & 0.961 & 0.923\\
        & F1  & 0.849 & 0.748 & 0.835& 0.781 & 0.878 & 0.878 & 0.881 & 0.796\\
        Torso & Accuracy  & 0.936 & 0.944 & 0.953& 0.952 & 0.960 & 0.960 & 0.947 & 0.852 \\
        & F1 & 0.233 & 0.102 & 0.007& 0.000 & 0.302 & 0.302 & 0.477 & 0.336\\
        Back & Accuracy & 0.942 & 0.939 & 0.935& 0.882 & 0.870 & 0.842 & 0.924 & 0.885\\
        & F1 & 0.241 & 0.000 & 0.401& 0.378 & 0.323 & 0.304 & 0.401 & 0.372\\
        Neck and/or Shoulders & Accuracy & 0.944& 0.945 & 0.940 & 0.943 & 0.940  & 0.944 & 0.931 & 0.944\\
        & F1 & 0.000 & 0.028 & 0.157& 0.095 & 0.083 & 0.000 & 0.077 & 0.000\\
        Arms & Accuracy & 0.867 & 0.874 & 0.886& 0.886 & 0.888 & 0.888 & 0.900 & 0.834\\
        & F1 & 0.403 & 0.423 & 0.392& 0.392 & 0.059 & 0.000 & 0.468 & 0.527\\
        Armpits & Accuracy & 0.971 & 0.971 & 0.967& 0.967 & 0.925 & 0.884 & 0.965 & 0.938\\
        & F1 & 0.000 & 0.000 & 0.221& 0.221 & 0.349 & 0.296 & 0.454 & 0.403\\
        Hands & Accuracy & 0.930 & 0.908 & 0.934& 0.870 & 0.975 & 0.975 & 0.971 & 0.966\\
        & F1 & 0.789 & 0.776 & 0.793& 0.711 & 0.923 & 0.926 & 0.931 & 0.916\\
        Legs & Accuracy & 0.875 & 0.767 & 0.866& 0.818 & 0.945 & 0.935 & 0.931 & 0.916\\
        & F1 & 0.338 & 0.535 & 0.498& 0.484 & 0.801 & 0.789 & 0.756 & 0.757\\
        Feet & Accuracy & 0.956 & 0.941 & 0.952& 0.918 & 0.981 & 0.979 & 0.976 & 0.975\\
        & F1 & 0.598 & 0.411 & 0.558& 0.591 & 0.869 & 0.859 & 0.832 & 0.840\\
        \bottomrule
    \end{tabular}
    \caption{Performance of our approach on a class-by-class basis on the medical images case study with CLIP image representations and with CLIP probability distributions.}
    \label{tab:images_class}
\end{table*}

\begin{table*}[t]
    \centering
    \begin{tabular}{l|c|cc|cc|cc|cc}
        \toprule
         \multirow{4}{*}{Metric} & \multirow{4}{*}{CLIP} & \multicolumn{8}{c}{Our Approach} \\
         & & \multicolumn{4}{|c|}{CLIP Image Representation} & \multicolumn{4}{c}{CLIP Probability Distribution} \\
         & & \multicolumn{2}{|c|}{$N_L = 59$} & \multicolumn{2}{|c|}{$N_L = 113$} & \multicolumn{2}{|c|}{$N_L = 54$} & \multicolumn{2}{|c}{$N_L = 116$} \\
         & & Majority Vote & Snorkel & Majority Vote & Snorkel & Majority Vote & Snorkel & Majority Vote & Snorkel \\
         \midrule
         Face & 0.891 & 0.708 & 0.774 & 0.712 & 0.636 & 0.574 & 0.602 & 0.606 & 0.606\\
         Scalp & 0.911 & 0.777 & 0.680 & 0.819 & 0.798 & 0.878 & 0.878 & 0.859 & 0.839\\
         Torso & 0.667 & 0.204 & 0.248 & 0.512 & 1.000 & 0.393 & 0.302 & 0.439 & 0.414\\
         Back & 0.000 & 0.349 & 0.420 & 0.333 & 0.345 & 0.331 & 0.308 & 0.393 & 0.384\\
         Neck and/or Shoulders & 0.167 & 0.220 & 0.500 & 0.261 & 0.368 & 0.131 & 0.083 & 0.173 & 0.104\\
         Arms & 0.750 & 0.350 & 0.314 & 0.402 & 0.446 & 0.152 & 0.074 & 0.489 & 0.465\\
         Armpits & 0.149 & 0.089 & 0.118 & 0.230 & 0.333 & 0.330 & 0.312 & 0.424 & 0.454\\
         Hands & 0.943 & 0.736 & 0.710 & 0.767 & 0.748 & 0.923 & 0.925 & 0.915 & 0.917\\
         Legs & 0.830 & 0.552 & 0.584 & 0.449 & 0.463 & 0.800 & 0.794 & 0.750 & 0.743\\
         Feet & 0.827 & 0.865 & 0.865 & 0.737 & 0.845 & 0.868 & 0.860 & 0.829 & 0.832\\
         \midrule
         Overall ACC & 0.552 & 0.551 & 0.565 & 0.584 & 0.604 & 0.636 & 0.634 & 0.673 & 0.669\\
         Weighted F1 & 0.527 & 0.552 & 0.540 & 0.578 & 0.576 & 0.633 & 0.619 & 0.672 & 0.661\\
         \bottomrule
    \end{tabular}
    \caption{Performance comparison (F1 score) between labels generated by our approach and the top-1 label output by CLIP}
    \label{tab:multiclass_performance}
\end{table*}

\subsection{Ablation Study}
\label{sec:ablation}

\begin{figure*}[t]
    \centering
    \begin{subfigure}[b]{0.45\textwidth}
       \includegraphics[width=\textwidth]{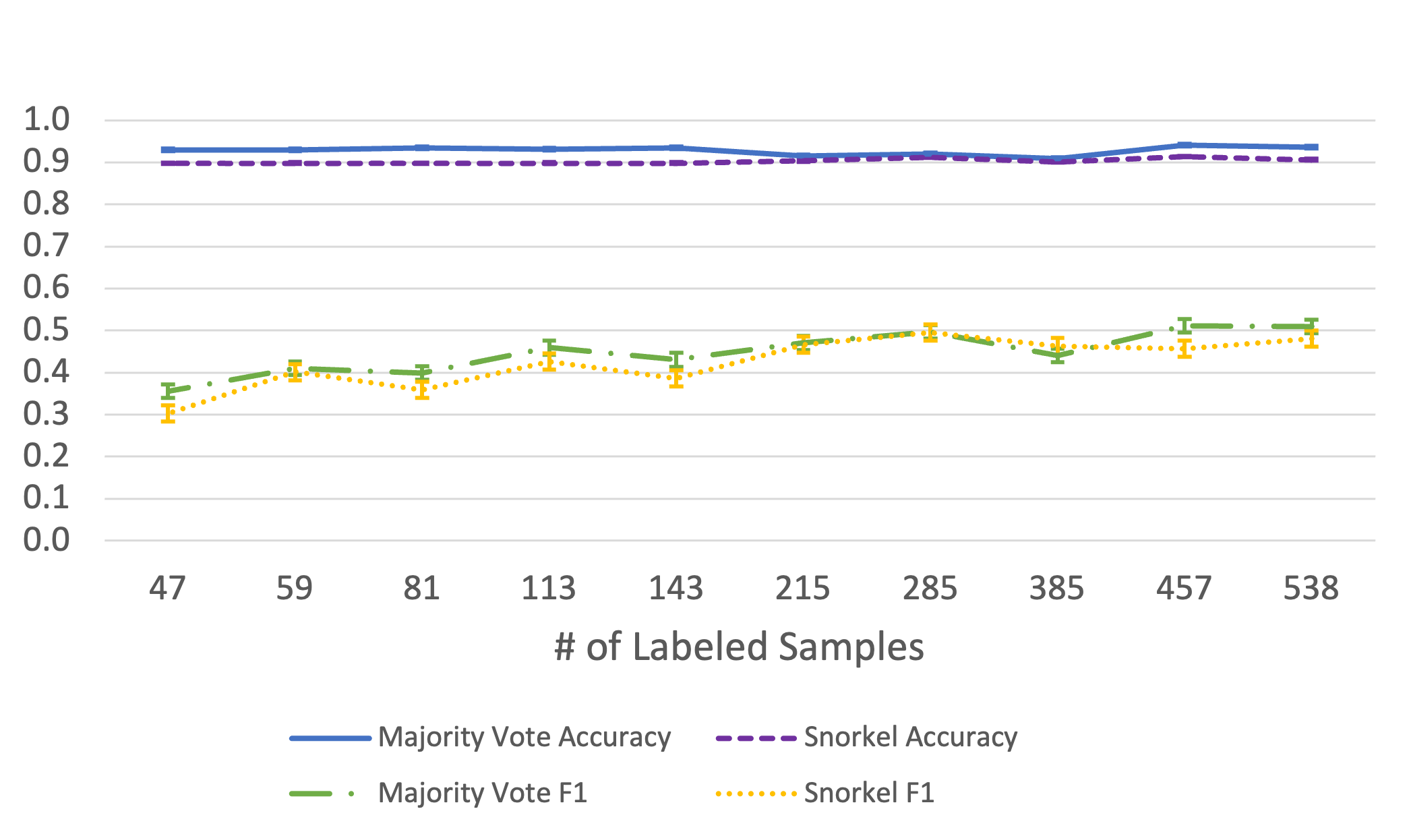}
       \caption{Euclidean distance using CLIP image representation (average over 10 binary classification for body part)}
       \label{fig:euc_binary} 
    \end{subfigure}
    \begin{subfigure}[b]{0.45\textwidth}
       \includegraphics[width=\textwidth]{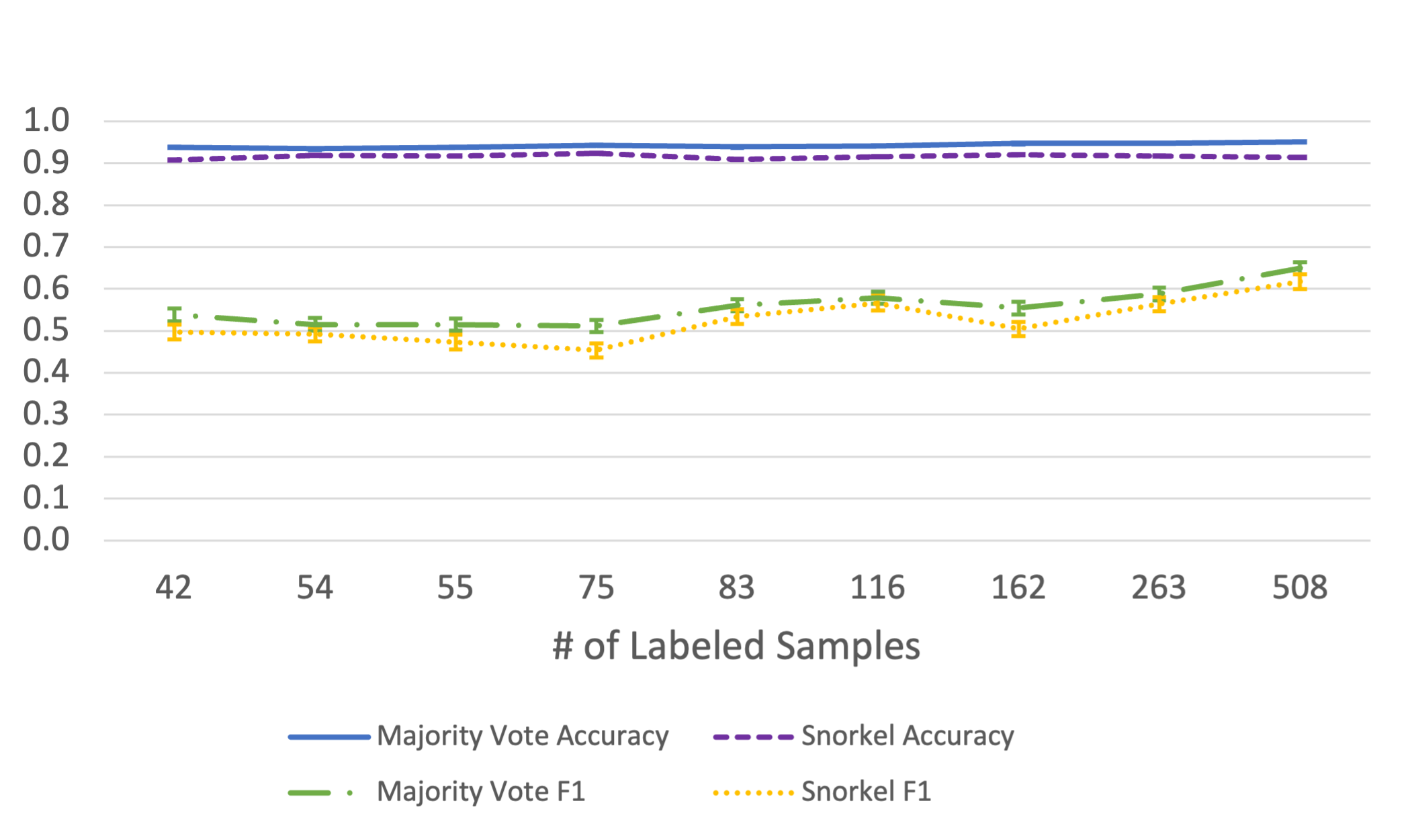}
       \caption{KL divergence using CLIP probability distribution (average over 10 binary classification for body part)}
       \label{fig:kl_binary} 
    \end{subfigure}
    \begin{subfigure}[b]{0.45\textwidth}
       \includegraphics[width=\textwidth]{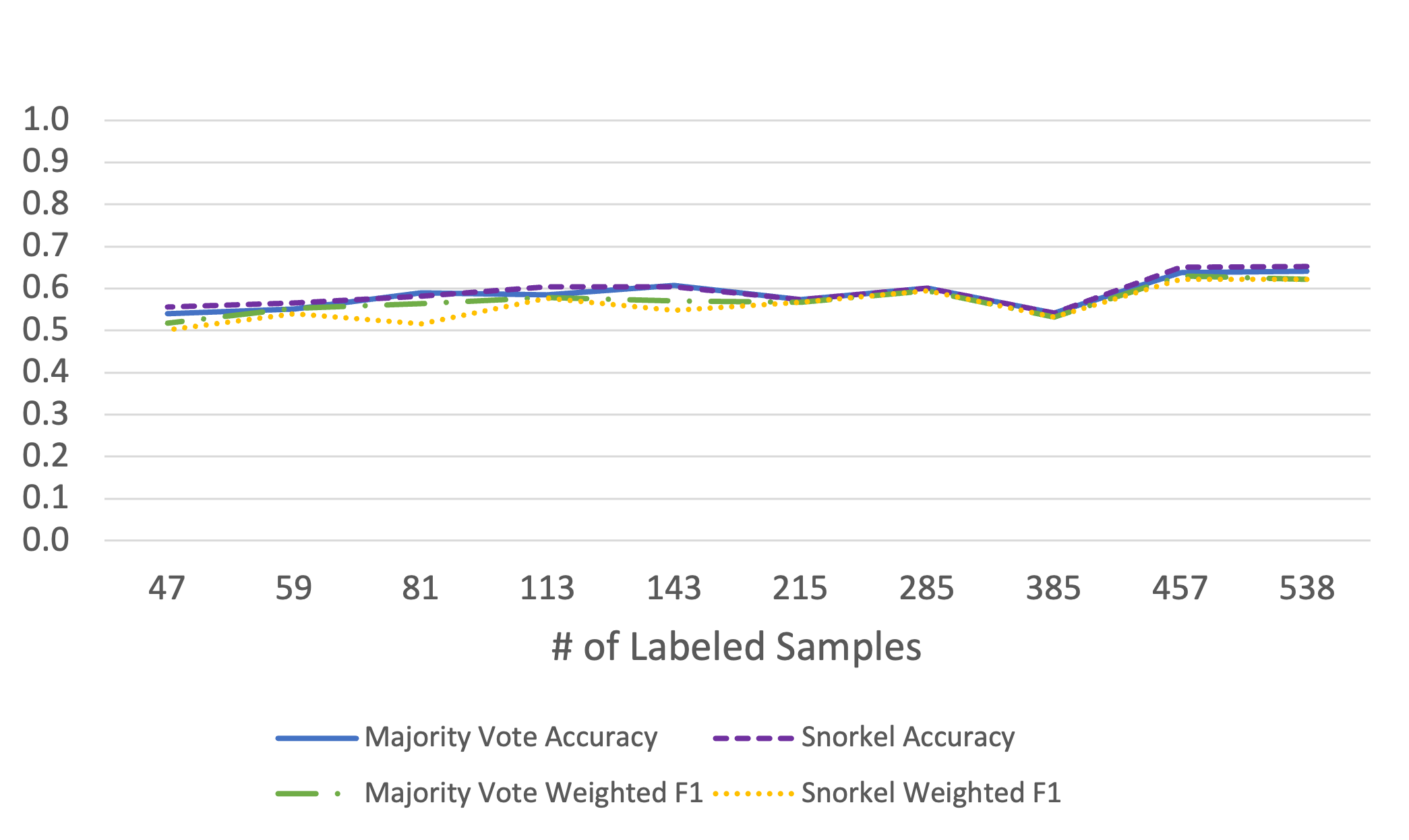}
       \caption{Euclidean distance using CLIP image representation (multi-class classification)}
       \label{fig:euc_multiclass} 
    \end{subfigure}
    \begin{subfigure}[b]{0.45\textwidth}
       \includegraphics[width=\textwidth]{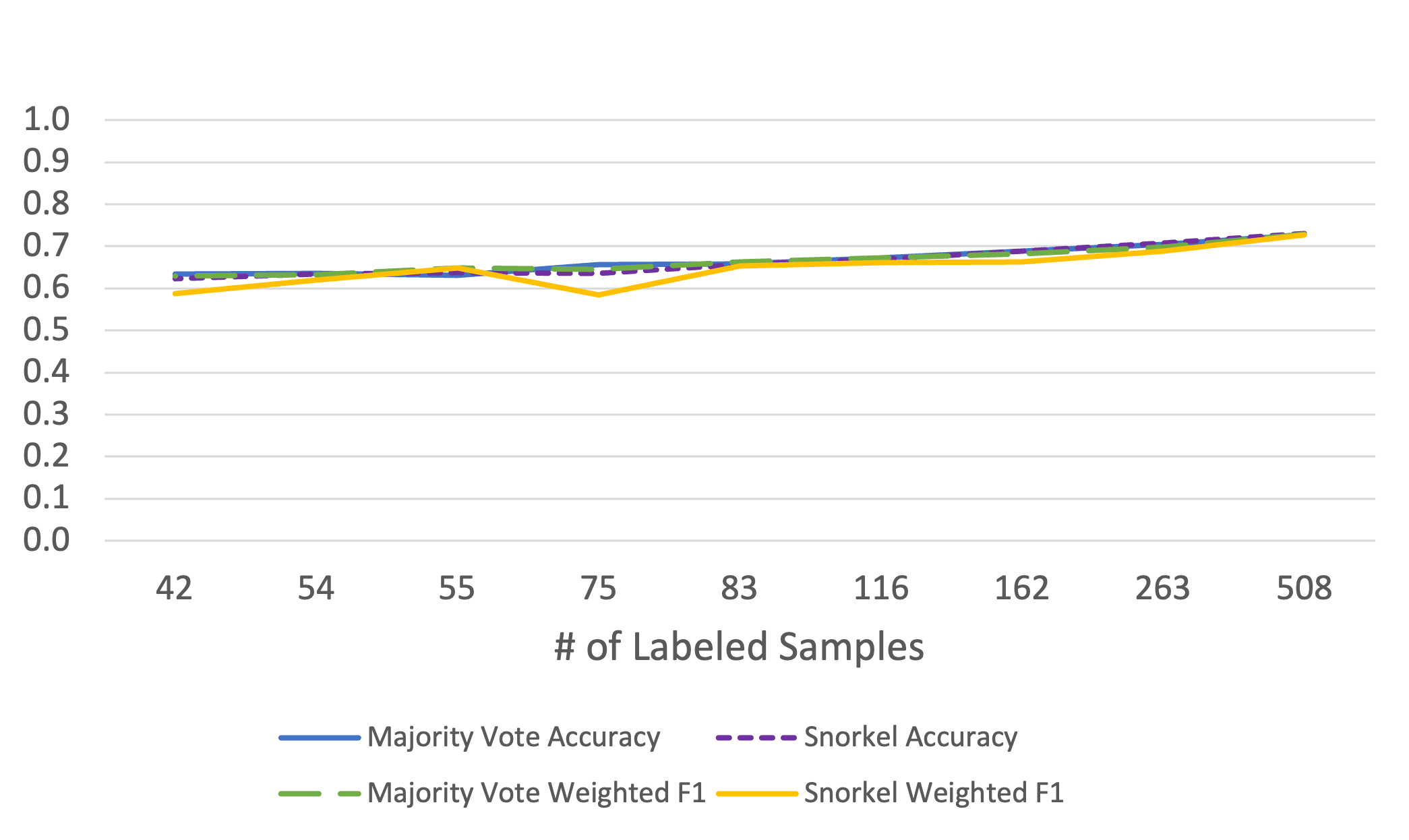}
       \caption{KL divergence using CLIP probability distribution (multi-class classification)}
       \label{fig:kl_multiclass} 
    \end{subfigure}
    \caption{F1 and Accuracy of our approach using varying number of labeled samples ($N_l$) for the medical image case study.}
    \label{fig:ablation_images}
\end{figure*}

\begin{figure}[t]
    \centering
    \includegraphics[width=0.45\textwidth]{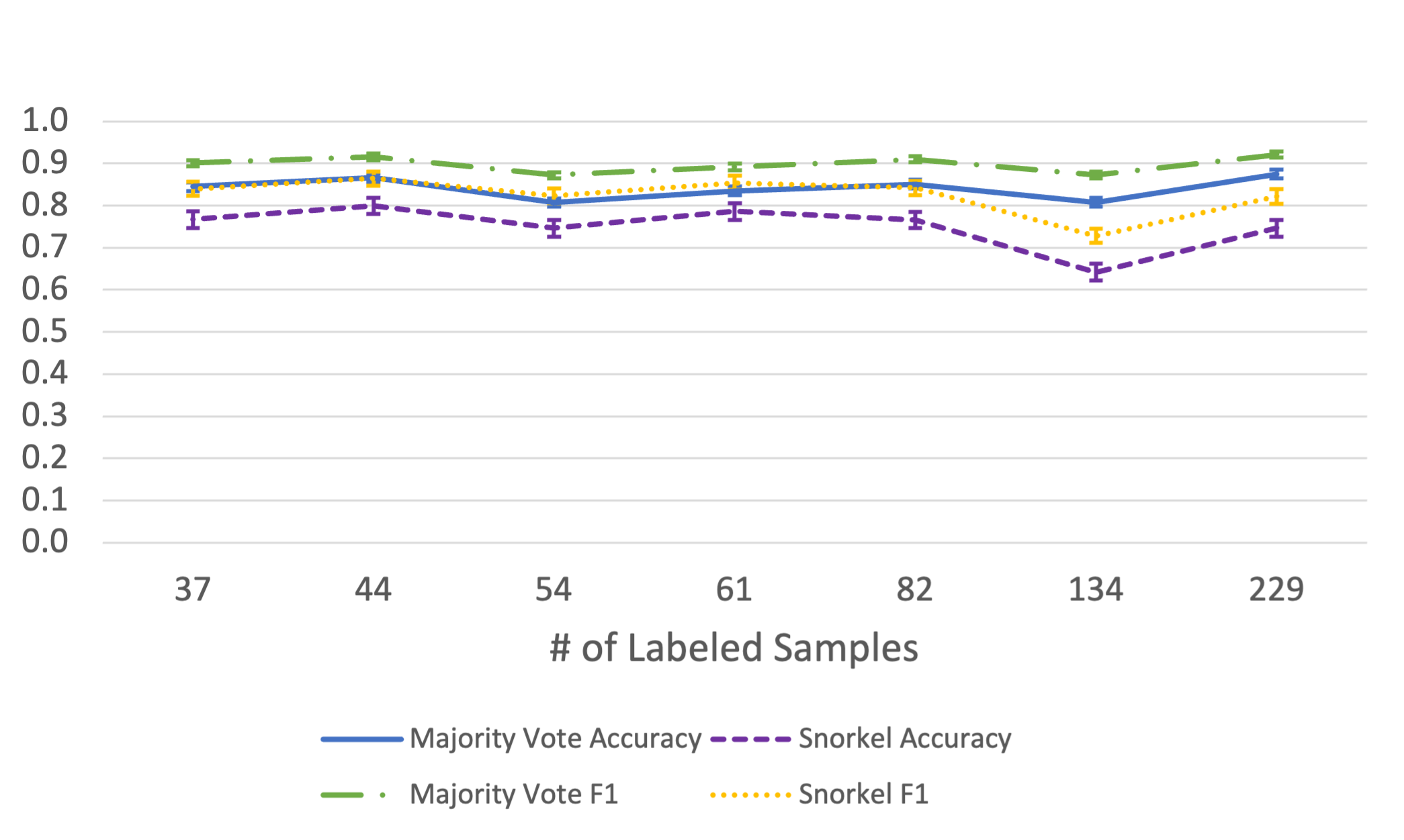}
    \caption{F1 and Accuracy of labels output by our approach using varying number of labeled samples ($N_L$) for the medical time series case study.}
    \label{fig:ablation_alarms}
\end{figure}

In this section, we analyze the impact of the number of labeled samples on the accuracy of labels produced by our approach. Note that the number of labeled samples used captures our labeling budget ($N_L$) when using a clinician in the loop. \revision{The value of $N_L$ depends on the distance threshold $t$, thus this ablation study directly investigates the influence of the distance threshold hyper-parameter on the performance of our approach.}\SP{TODO13 Adding this to clarify this ablation study explores the impact of the distance threshold hyperparam.} We plot the trend in labeling accuracy and F1 score for the time series case study in Figure~\ref{fig:ablation_alarms} and the medical image case study in Figure~\ref{fig:ablation_images}. In general, our approach demonstrates consistently high labeling accuracy and growing F1 as the number of labeled samples increases. \revision{However, we note that for the time series case study in Figure~\ref{fig:ablation_alarms}, in particular for Snorkel, the trend in label quality is noisy. We attribute this to the inherent noise in our approach and plan to quantify its impact on our approach's performance in future work.}\SP{TODO14}

\subsection{Limitations}
\revision{In this section, we discuss the limitations of our approach. Recall that our method relies on pre-selected distance functions and, for image data, high-dimensional features over which distance is computed. These items effect the performance of the memory generation algorithm which partitions the data and determines the subset of representative samples (memories) for the clinician to label. However, in certain real-world scenarios, the features and distance functions our approach uses may not be effective. The features may lack relevance for comparing samples or the distance functions may not effectively capture the similarity between samples. This can lead to a poor partitioning of the data, resulting in low accuracy weak labels when the clinician's labels for the memories are imposed on the respective partitions. Therefore, the effectiveness of our approach hinges heavily on the suitability of the pre-selected distance functions or high-dimensional features for the application.\SP{TODO6}}

\revision{As studied in Section~\ref{sec:ablation}, the performance of our approach is impacted by the clinician's labeling budget $N_L$. As a clinician increases their budget, the quality of the labels produced by our approach generally improves. Consequently, poorly chosen (\eg very small) labeling budgets can cause our approach to produce inaccurate probabilistic labels. For the evaluation in this paper, we explicitly chose labeling budgets that would be considered reasonable for an clinician in our case studies. However, in practice, determining an appropriate labeling budget may not be straightforward and may require consideration of several factors. For example, the complexity of the labeling task (\eg the time and expense associated with annotating a data sample) is an important factor for an clinician with time constraints. Additionally, a clinician may have a desired tolerance for incorrect labels (\eg a threshold on false positive labels), thus posing a trade-off where a larger labeling budget may be necessary to ensure higher label quality. We plan to explore these factors and devise guidelines based on them for selecting the labeling budget $N_L$ in future work.\SP{TODO7}}

\revision{Lastly, our approach requires the clinician to manually label a subset of the data determined by the memory generation algorithm, with the subset's size being bounded by the clinician's labeling budget. As mentioned previously, manual labeling can be both time-consuming an expensive. Therefore, despite being a small portion of the data, this requirement can still pose a limitation in certain real-world scenarios.\SP{TODO9} In the medical time series data case study, approximately 54 seconds was spent labeling each alarm in the dataset, roughly 55 hours to label all 3,625 alarms~\cite{macmurchy2017acceptability}. Using our approach with a labeling budget of 134 alarms, a clinician would spend only 2 hours labeling.\SP{TODO10} 
In the medical images case study, we demonstrate that our approach can produce high quality labels and only required an expert to label approximately 100 images instead of thousands, reducing the number of samples to be manually labeled by roughly 99\%.
\SP{TODO11} Hence, in both case studies presented in this paper, our approach demonstrates its capability to significantly reduce manual labeling efforts while still achieving high-quality labels.}
\section{Conclusion}

Labeling function generation is a challenging problem in the data-programming paradigm. An assumption made when automatically coming up with weak-labeling functions is that a small labeled set is available. In this paper we show that carefully selecting \emph{which} samples to label and querying an expert clinician post facto can have a significant impact on performance. Additionally when compared to similar approaches this method does not need any classifier training to construct the weak labeling function from the labeled samples. In the future our plan is to \revision{study how the memory generation algorithm optimization hyper-parameters impact the accuracy of labels produced by our approach.\SP{TODO13} Furthermore, we plan to} reduce the number of labeled samples required, thereby reducing the load on the clinician. We believe that leveraging large language models and existing medical guidelines may enable such an improvement. 

\revision{
\section*{Acknowledgment}
This work was supported in part by NSF-1915398 and ARO MURI W911NF-20-1-0080.
}\SP{TODO5}

\bibliographystyle{IEEEtran}
\bibliography{IEEEabrv,references}

\end{document}